\documentclass[10pt,conference,a4paper]{IEEEtran}

\usepackage{cite}
\ifCLASSINFOpdf
\usepackage[pdftex]{graphicx}
\else
\fi
\usepackage{amsmath}
\usepackage{array}
\usepackage{url}
\graphicspath{ {images112/} }
\begin{document}
	%
\title{How do Convolutional Neural Networks \protect\\ Learn Design?}
\author{
	\IEEEauthorblockN{
		Shailza Jolly\IEEEauthorrefmark{1},
		Brian Kenji Iwana\IEEEauthorrefmark{2},
		Ryohei Kuroki\IEEEauthorrefmark{2},
		Seiichi Uchida\IEEEauthorrefmark{2}
		}
	\IEEEauthorblockA{
		\IEEEauthorrefmark{1}University of Kaiserslautern, Kaiserlautern, Germany\\ Email: sjolly@rhrk.uni-kl.de
		}
	\IEEEauthorblockA{
		\IEEEauthorrefmark{2}Department of Advanced Information Technology, Kyushu University, Fukuoka, Japan\\ Email: \{brian, kuroki, uchida\}@human.ait.kyushu-u.ac.jp
		}
		
	}
\maketitle
\begin{abstract}
 
In this paper, we aim to understand the design principles in book cover images which are carefully crafted by experts. Book covers are designed in a unique way, specific to genres which convey important information to their readers. By using Convolutional Neural Networks~(CNN) to predict book genres from cover images, visual cues which distinguish genres can be highlighted and analyzed. In order to understand these visual clues contributing towards the decision of a genre, we present the application of Layer-wise Relevance Propagation~(LRP) on the book cover image classification results. We use LRP to explain the pixel-wise contributions of book cover design and highlight the design elements contributing towards particular genres. In addition, with the use of state-of-the-art object and text detection methods, insights about genre-specific book cover designs are discovered.
\end{abstract}
	%
\IEEEpeerreviewmaketitle
\section{Introduction}
Visual design renders specific impressions to transmit information which enriches the product's value. However, these visual designs despite of being important are not analyzed objectively or statistically. Analyzing these visual designs enables us to understand the contained information carried by them.

An interesting target of visual design analysis is book cover image design where the design of a book cover can infer the genre. 
Each book cover is carefully designed by typographers and their designs represent the book contents in an intuitive way for better sales. This association of books to specific genres is based on the differences in their underlying book cover designs~\cite{iwana2016judging}. The slight change in book cover design can reflect changes in book genre which makes design learning a challenging task for book covers.

In order to understand the design elements used for machine aided book cover classification, we employ Convolutional Neural Networks~(CNN)~\cite{lecun1998gradient}.
In recent years, CNNs have achieved state-of-the-art results in isolated character recognition~\cite{ciregan2012multi,uchida2016further} and large-scale image recognition~\cite{simonyan2014very,szegedy2015going}.
Notably, Iwana~et~al.~\cite{iwana2016judging} demonstrated that CNNs can be used for genre classification based on book cover image, although with a high level of difficulty. 
However, that study was subjective and not enough explanation is given as to why the CNN performed as it did.
	
To interpret the reasoning behind a CNN's prediction we used a method called Layer-wise Relevance Propagation~(LRP) ~\cite{bach2015pixel}. LRP decomposes output function on its input variables and highlights input pixels contributing towards the network decision. It produces a layer-wise relevance heatmap by recursively multiplying the relevance of higher layers by the normalized feature maps of the target layer. 
The heatmaps can help us to discover the input image elements which have an effect on the classification result. 

The main contributions of this paper are threefold.
Firstly, we classified the book cover images using one-vs-others classification with CNNs.
Secondly, the models built by the CNNs are analyzed using LRP.
With LRP, we demonstrate design elements specifically relevant to classification of the book cover images. We show that certain objects have a strong relevance to particular genres. 
Finally, we use state-of-the-art object detection and text detection methods, namely Single Shot Multibox Detector~(SSD)~\cite{liu2016ssd} and Efficient and Accurate Scene Text Detector~\cite{zhou2017east}, to quantitatively enforce the results found by LRP. 
This reveals the specific elements in which CNNs classify book cover images for genre classification. 
	
The organization is as follows. Section~\ref{rel} provides related works in design understanding and genre classification as well as feature visualization of CNNs. Section~\ref{cnn} reviews the 
data and tools used for understanding book cover design.
Section~\ref{lrpt} presents analysis of CNN's understanding of book cover design. In Section~\ref{ssd}, we demonstrate the use of LRP combined with SSD and EAST for quantitative analysis. Finally, Section~\ref{con} draws a conclusion.
	
\section{Related Work}
\label{rel}
\subsection{Genre Classification}
Artistic style understanding and subjective genre classification is a budding field in machine learning. 
For example, recent attempts have been done to identify artistic styles and quality of paintings and photographs~\cite{datta2008image,karayev2013recognizing} with neural network models. 
In addition, there have been trials to classify music by genre~\cite{tzanetakis2002musical,mckay2004automatic}, book covers by genre~\cite{iwana2016judging}, movie posters by genre~\cite{chu2017movie}, paintings by genre~\cite{zujovic2009classifying}, and text by genre~\cite{finn2006learning,petrenz2011stable}.
Also, in a general sense, document classification can be considered genre classification and deep CNNs are the state-of-the-art in the document classification domain~\cite{kang2014convolutional,harley2015evaluation,afzal2015deepdocclassifier}.

\subsection{Visualization inside of CNNs}
There is a desire to visualize features and determine pixel-wise attention and relevance within the hidden layers of CNNs. 
However, this is a not a straightforward task~\cite{erhan2009visualizing}. 
Erhan et al.~\cite{erhan2009visualizing} proposed using gradient decent to maximize a node's activation to visualize the employed features. 
Similar work has been done for large-scale image classification~\cite{olah2017feature}.
Zeiler and Fergus~\cite{zeiler2014visualizing} used deconvolutional neural networks to visualize features learned by CNNs. In addition, they created heatmaps by monitoring class changes systematic cover up of portions of the images.  
Class Activation Maps~(CAM)~\cite{zhou2016learning}, GradCAM~\cite{selvaraju2016grad}, and GradCAM++~\cite{chattopadhyay2017grad} reveal the parts of images which are most important to a class using global average pooling~(GAP). 
	
Recently, LRP has been used in the fields of text~\cite{arras-plos17} where classification scores were projected back to input features for extracting relevant words for a specific prediction. The method has also shown successes in model understanding in fields of sentiment analysis~\cite{ArrWASSA17}, action recognition~\cite{SriICASSP17}, 
and age and gender classification~\cite{arbabzadah-gcpr16}. 
As far as the authors are aware, this is the first time LRP has been used for the understanding of genre or design classification.

\section{Data and tools for understanding book cover design}
\label{cnn}
\subsection{Amazon Book Cover Dataset}
We used the \textit{Book Cover Image to Genre} dataset\footnote{\url{https://github.com/uchidalab/book-dataset}} Task 1.A. 
The dataset consists of 57,000 book cover images divided into 30 classes of equal sizes. 
In the experiments, we used the predefined training set and test set modified for one-vs-others classification. 
In this way, genre-wise training sets were prepared with an equal distribution of positive and negative data samples.

\subsection{Convolution Neural Networks}
CNNs are able to tackle image recognition by implementing convolutions of learned filter-like shared weights which maintain the structural qualities of images while acting as feature extractors~\cite{lecun1998gradient}. For the experiment, we implement CNNs to tackle book genre classification. 
To use the book cover images with a CNN, they were preprocessed by scaling them to 112$\times$112 pixels by 3 color channel images and by normalizing the values to be between -1 and 1. 
The CNN used for the experiments has six convolutional layers with Rectified Linear Units~(ReLU) activations and a softmax output layer. 
The convolutional layers consisted of three layers of 10 nodes with $5\times5$ convolutional filters, one layer of 25 nodes with a $4\times4$ filter, one layer of 50 nodes with a $3\times3$ filter, and one layer of 100 nodes with a $1\times1$ filter. 
A $2\times2$ maxpooling layer with stride 2 was used between each convolution layer.
Finally, the CNNs were trained using gradient decent with a batch size of 25 at a learning rate of 0.001 for 50,000 iterations.

\begin{figure}
\centering
\includegraphics[width=.80\columnwidth]{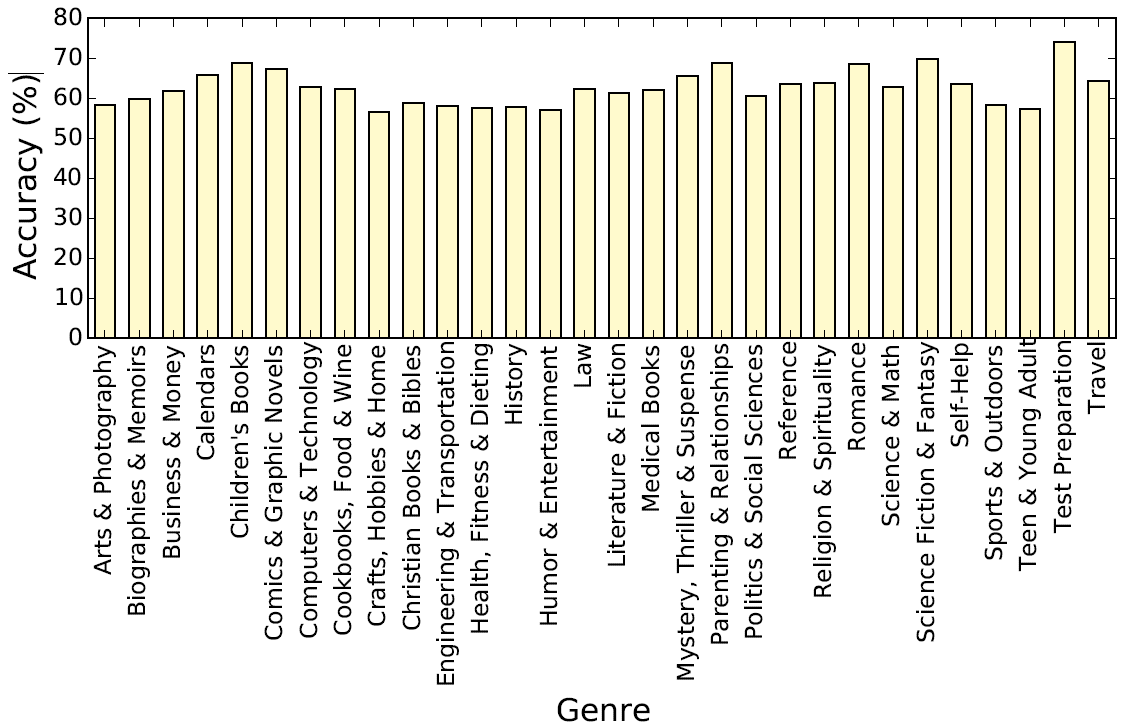}
\caption{\label{fig:table} CNN accuracy by genre.}
\end{figure}

The accuracy results for each genre is summarized in Fig.~\ref{fig:table}. 
In particular, the CNNs had difficulties with the reference classes, such as "Engineering \& Transportation," "Health, Fitness and Dieting," "History," "Medical Books," and "Reference." 	Conversely, "Children Books," "Romance," and "Test Preparation" had high accuracies. 
However, more than just classification accuracy, the purpose of this paper is to understand why the CNN's performed as such and reveal the relevant parts of the images.

\subsection{Layer-wise Relevance Propagation}
The LRP algorithm and the LRP toolbox~\cite{lapuschkin2016lrp} aims to explain the reasoning behind the decision made by a network model which allows its users to validate classifier results. LRP is mainly derived from Deep Taylor Decomposition~\cite{montavon2017explaining}, a method of decomposing network's output predictions onto its input variable. The results after such a decomposition is visualized in the form of a heatmap highlighting each pixel's importance for the prediction. 

\begin{figure}
\begin{center}
\setlength\fboxsep{0pt}
\setlength\fboxrule{0pt}
\framebox[0.48\columnwidth]{\includegraphics[clip, trim=1cm 19.25cm 8cm 1cm, width=0.45\columnwidth]{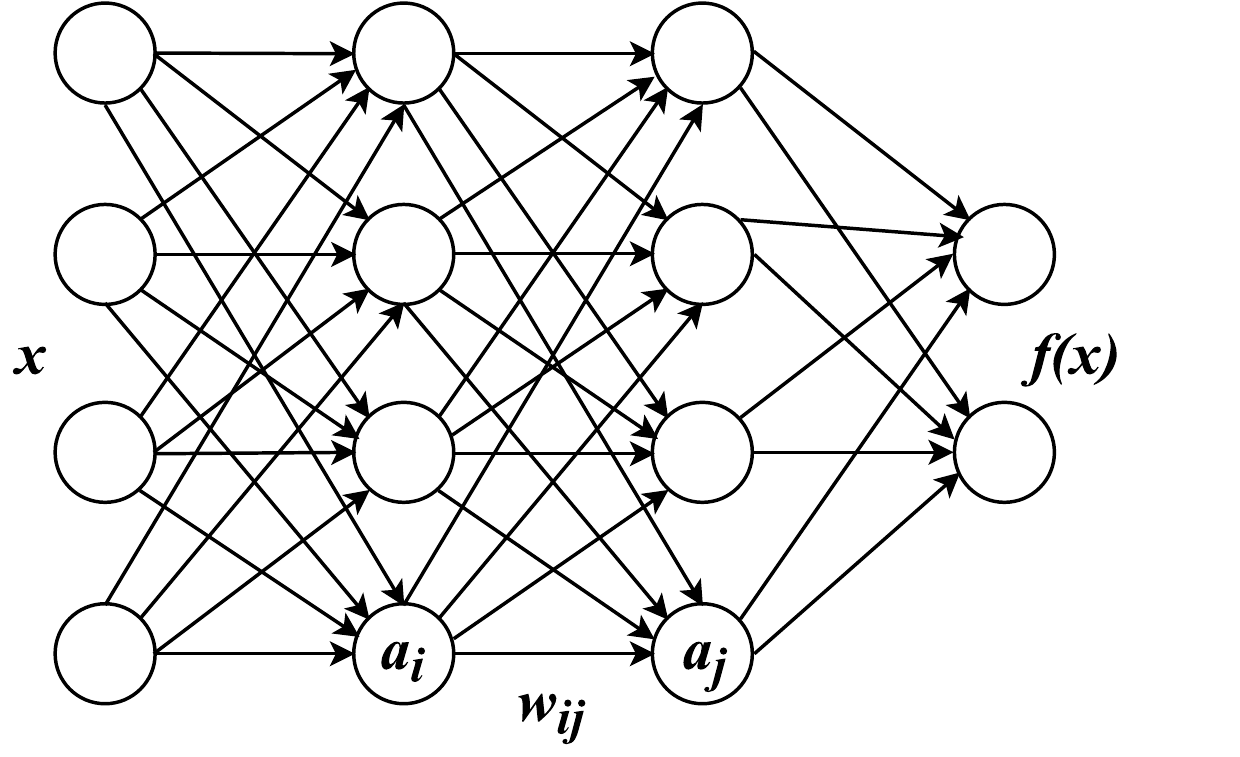}}
\framebox[0.48\columnwidth]{\includegraphics[clip, trim=1cm 19.25cm 8cm 1cm, width=0.45\columnwidth]{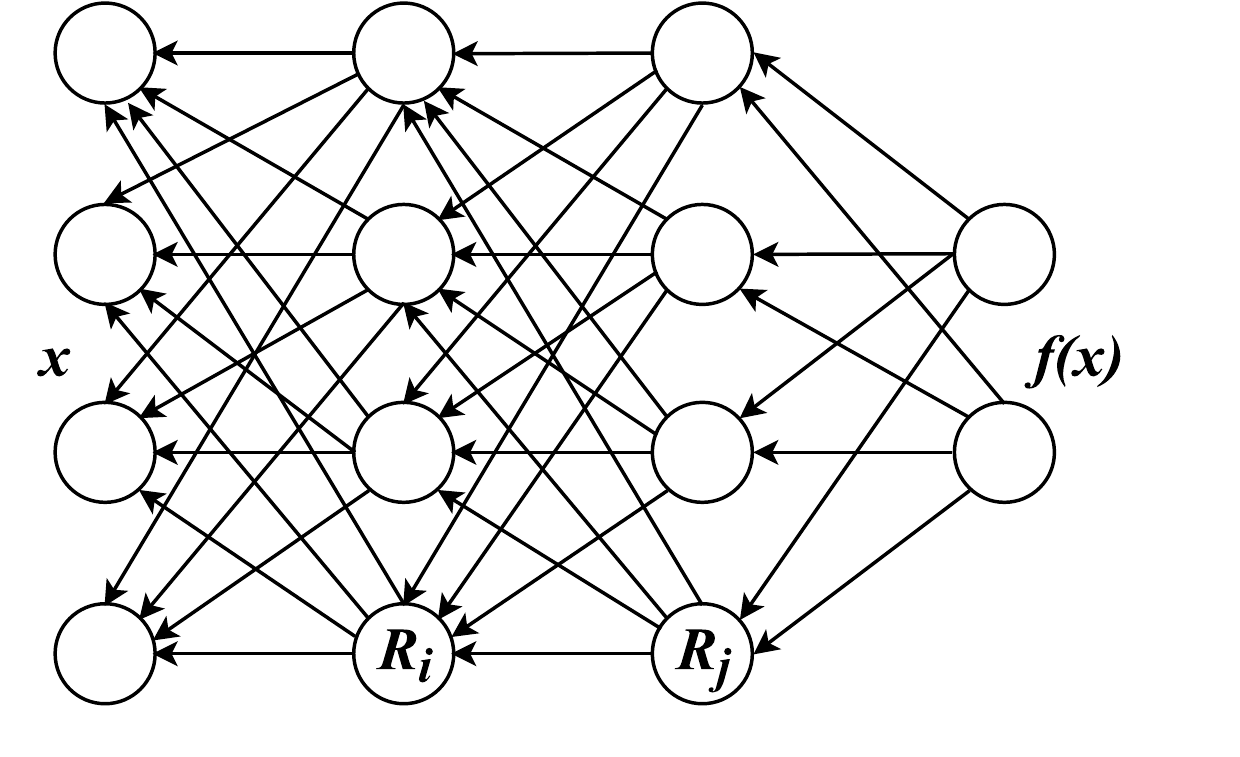}}			
\end{center}
\caption{Feed forward neural network with the (left) forward pass and the (right) backward relevance calculation. The function $f(x)$ is the prediction outcome given input $x$. The variables $a_i$ and $a_j$ are the inputs for node $i$ and $j$, respectively. $R_i$ is the relevance of node $i$ and $R_j$ is the relevance of node $j$.}
\label{lrp}
\end{figure}

LRP explains output function, i.e. classifier's decision, which helps us to derive all of the crucial pixels for a particular prediction. In Fig.~\ref{lrp}, the technique is shown in which the output value given by the network is decomposed backwards layer by layer until it reaches the input. This backward decomposition of network's prediction uses local redistribution rules for assigning relevance values $R_i$ to each neuron contributing towards the output, namely
\begin{equation}
\sum_{i} R_i = \sum_{j} R_j =\dots= \sum_{k}R_k = f(x),
\label{cons}
\end{equation}
where $f(x)$ is the prediction function, $R_i$ is the relevance of node $i$ in the target layer, $R_j$ is the relevance of node $j$ of the previous layer, and $R_k$ is the relevance of node $k$ of the highest layer. 
The total amount of relevance is conserved in this equation. 

For the experiment, we used the $\alpha-\beta$ decomposition formula defined by
\begin{equation}
	R_i = \sum_{j} \left(\alpha \frac{(a_i w_{ij})^+}{\sum_{i} ( a_i w_{ij}) ^ +} + \beta  \frac{(a_i w_{ij})^-}{\sum_{i} ( a_i w_{ij}) ^ -}   \right) R_j,
\label{alpha}
\end{equation}  
where $\alpha$ and $\beta$ are hyperparameters to weight the positive values of $\frac{(a_i w_{ij})}{\sum_{i} ( a_i w_{ij})}$ and the negative values of $\frac{(a_i w_{ij})}{\sum_{i} ( a_i w_{ij})}$, respectively. 
Furthermore, $w_{ij}$ is the weight between nodes $i$ and $j$ and $a_i$ is the input to node $i$. 
This decomposition allows for the separation of the positive connections and the negative connections. 
Values inside positive bracket indicates propagation of activating input messages while negative weight connections indicate deactivating input values.

\subsection{Single-Shot Multibox Detector}
To develop a better understanding of the objects within book cover images, we employed SSD~\cite{liu2016ssd}, a state-of-the-art deep neural network based object detection method. 
SSD is a feed forward CNN which produces a multi-scale collection of fixed size bounding boxes and scores for object detection within the boxes. 
A final non-maximal suppression step determines the final detections. 
The result of SSD is bounding box regions with object classification labels. 
Using SSD, it is possible to accurately detect multiple objects of different classes within images.

\subsection{Efficient and Accurate Scene Text Detector}
For humans, text is an important component of book covers; it is where the title, authors, and additional information is conveyed. 
However, a CNN may place a different importance on text than humans. 
Thus, to analyze the relevance of text in book covers, we use EAST~\cite{zhou2017east} as a text detector. 
EAST uses a multi-channel Fully Convolutional Network~(FCN) and non-maximal suppression on predicted geometric shapes to detect multi-orient text-line and word boxes.

	
\section{How CNNs Understand Book Cover Design: Qualitative Analysis}
\label{lrpt}
In this section, we have presented LRP results from main genres. The analysis helped us to deduce book cover design elements contributing towards a prediction by CNN. 
We used $\alpha-\beta$ decomposition formula with values of $\alpha=2$ and $\beta=-1$ which is suggested for networks using ReLU activation functions because it emphasizes the positive elements and de-emphasizes the negative ones~\cite{bach2015pixel}. 
This is important due to the ReLU activation function setting negative values to zero. 
In the heatmaps generated by LRP under this decomposition, pixels adding positive contribution are represented in red color and the ones adding negative contribution are represented by blue color. 
	
\subsection{Sports \& Outdoors}
\begin{figure*}
\begin{center}
\setlength\fboxsep{0pt}
\setlength\fboxrule{0pt}
\framebox[0.30\columnwidth]{\includegraphics[width=0.30\columnwidth]{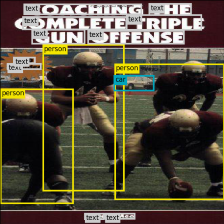}}
\framebox[0.30\columnwidth]{\includegraphics[width=0.30\columnwidth]{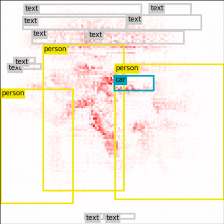}}
\framebox[0.08\columnwidth]{\rule{.1pt}{2.7cm}}
\framebox[0.30\columnwidth]{\includegraphics[width=0.30\columnwidth]{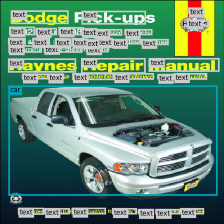}}
\framebox[0.30\columnwidth]{\includegraphics[width=0.30\columnwidth]{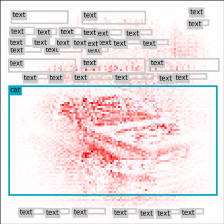}}
\framebox[0.08\columnwidth]{\rule{.1pt}{2.7cm}}
\framebox[0.30\columnwidth]{\includegraphics[width=0.30\columnwidth]{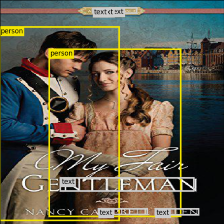}}
\framebox[0.30\columnwidth]{\includegraphics[width=0.30\columnwidth]{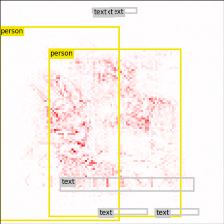}}

\framebox[0.30\columnwidth]{\includegraphics[width=0.30\columnwidth]{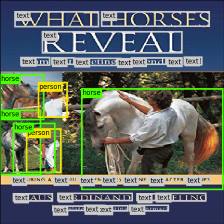}}
\framebox[0.30\columnwidth]{\includegraphics[width=0.30\columnwidth]{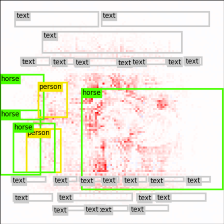}}
\framebox[0.08\columnwidth]{\rule{.1pt}{2.8cm}}
\framebox[0.30\columnwidth]{\includegraphics[width=0.30\columnwidth]{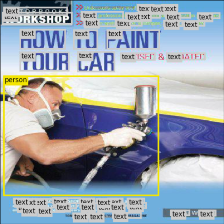}}	
\framebox[0.30\columnwidth]{\includegraphics[width=0.30\columnwidth]{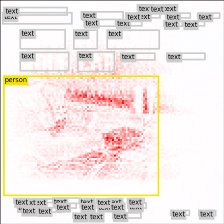}}
\framebox[0.08\columnwidth]{\rule{.1pt}{2.8cm}}
\framebox[0.30\columnwidth]{\includegraphics[width=0.30\columnwidth]{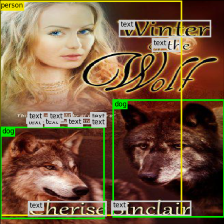}}
\framebox[0.30\columnwidth]{\includegraphics[width=0.30\columnwidth]{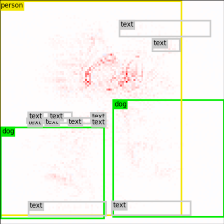}}
\framebox[0.60\columnwidth]{(a) ``Sports \& Outdoors''}
\framebox[0.08\columnwidth]{}
\framebox[0.60\columnwidth]{(b) ``Engineering \& Transportation''}
\framebox[0.08\columnwidth]{}
\framebox[0.60\columnwidth]{(c) ``Romance''}
\end{center}

\begin{center}
\setlength\fboxsep{0pt}
\setlength\fboxrule{0pt}
\framebox[0.30\columnwidth]{\includegraphics[width=0.30\columnwidth]{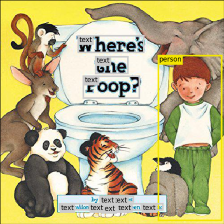}}
\framebox[0.30\columnwidth]{\includegraphics[width=0.30\columnwidth]{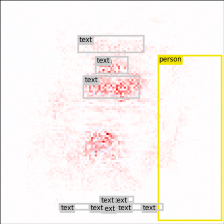}}
\framebox[0.08\columnwidth]{\rule{.1pt}{2.7cm}}
\framebox[0.30\columnwidth]{\includegraphics[width=0.30\columnwidth]{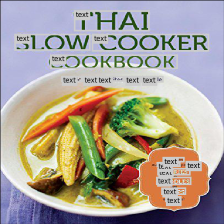}}
\framebox[0.30\columnwidth]{\includegraphics[width=0.30\columnwidth]{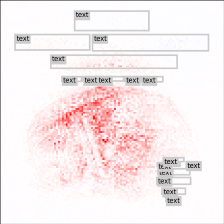}}
\framebox[0.08\columnwidth]{\rule{.1pt}{2.7cm}}
\framebox[0.30\columnwidth]{\includegraphics[width=0.30\columnwidth]{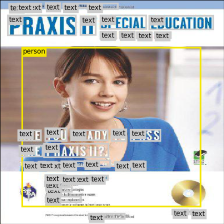}}
\framebox[0.30\columnwidth]{\includegraphics[width=0.30\columnwidth]{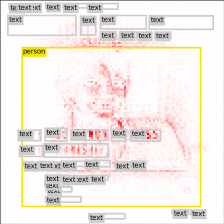}}

\framebox[0.30\columnwidth]{\includegraphics[width=0.30\columnwidth]{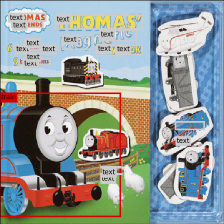}}
\framebox[0.30\columnwidth]{\includegraphics[width=0.30\columnwidth]{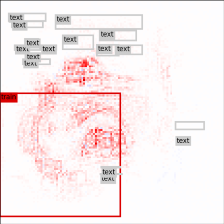}}
\framebox[0.08\columnwidth]{\rule{.1pt}{2.8cm}}
\framebox[0.30\columnwidth]{\includegraphics[width=0.30\columnwidth]{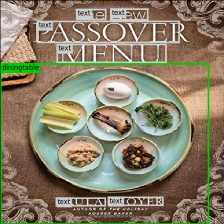}}
\framebox[0.30\columnwidth]{\includegraphics[width=0.30\columnwidth]{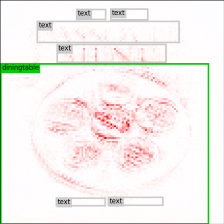}}
\framebox[0.08\columnwidth]{\rule{.1pt}{2.8cm}}
\framebox[0.30\columnwidth]{\includegraphics[width=0.30\columnwidth]{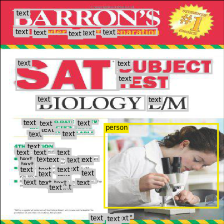}}
\framebox[0.30\columnwidth]{\includegraphics[width=0.30\columnwidth]{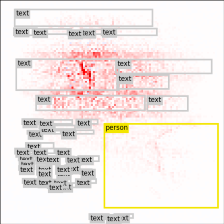}}

\framebox[0.60\columnwidth]{(d) ``Children's Books''}
\framebox[0.08\columnwidth]{}
\framebox[0.60\columnwidth]{(e) ``Cookbooks, Food \& Wine''}
\framebox[0.08\columnwidth]{}
\framebox[0.60\columnwidth]{(f) ``Test Preparation''}
\end{center}
\caption{\label{bookcovers} Correctly recognized book covers. Object classes by SSD and text by EAST are highlighted.}
\end{figure*}

Under this genre, many book covers with pictures of players playing indoor and outdoor games were seen. Figure~\ref{bookcovers}~(a) shows LRP results on these covers, which presents significance of player's picture on the cover.
The first image in Fig.~\ref{bookcovers}~(a) supports this fact with LRP being centered on players who are either playing a sport or showing some player like gesture, with car in background adding no contribution. The second image in Fig.~\ref{bookcovers}~(a) emphasizes the animal's importance for this genre's prediction.  

\subsection{Engineering \& Transportation}
For this genre, almost all the covers with vehicle pictures on their covers were classified correctly by the network. With LRP in Fig.~\ref{bookcovers}~(b), part of image containing cars or motorbikes seem to add more relevance than others. The last image in the Fig.~\ref{bookcovers}~(b) presents the cases when contribution of person image was dominated by vehicle in the image. 

\subsection{Romance}
Its obvious from the genre name that pictures of couples on the cover are going to have more relevance and LRP results showed this fact to be true. However, among pictures presented in Fig.~\ref{bookcovers}~(c), LRP depicted girls to add more relevance than men or other things. The reason could reside in their physical appearance, hairs, and choice of dresses. The same was demonstrated in last picture of Fig.~\ref{bookcovers}~(c) in which girl's hair are seen to add more relevance with zero relevance coming from animal part on book cover.

\subsection{Children's Books}
Almost all the children book covers contain pictures of cartoon characters. LRP on covers from this genre showed these cartoon characters to have higher relevance. An interesting result is shown in first picture of Fig.~\ref{bookcovers}~(d), where person is depicted as an adversarial identity and importance of cartoons in cover is highlighted. Some covers showed more relevance for one object in the set of objects. Like, in Fig.~\ref{bookcovers}~(d) some cartoons in last picture have higher relevance. It can be because of the object placement and their orientations. With the help of this information, one can make smart choices for different characters, cartoons and color patterns.
	

\subsection{Cookbooks, Food \& Wine Books}
Book covers in this genre most commonly contained pictures of different kinds of food. The results in Fig.~\ref{bookcovers}~(e) showed these food pictures as containment of higher relevance for this genre. However, carefully analyzing LRP results we discovered shapes of dishes like bowls or spoons adding significant relevance for the genre's prediction. So, this marks significance of dish shape designs on covers from this genre.

\subsection{Test Preparation}
\label{sec:testprep}
The genre contained covers with both text and pictorial information as shown in Fig.~\ref{bookcovers}~(f). With most contribution coming from big text content on covers. Images of Fig.~\ref{bookcovers}~(f) presents big texts to add more relevance than images of people. In first image of Fig.~\ref{bookcovers}~(f), despite of big girl face, relevance is concentrated on text area of book cover.

\begin{figure}
\begin{center}
			
\setlength\fboxsep{0pt}
\setlength\fboxrule{0pt}
\framebox[0.22\columnwidth]{\includegraphics[width=0.22\columnwidth]{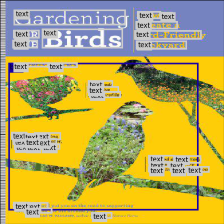}}
\framebox[0.22\columnwidth]{\includegraphics[width=0.22\columnwidth]{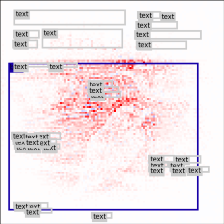}}
\framebox[0.06\columnwidth]{ }
\framebox[0.22\columnwidth]{\includegraphics[width=0.22\columnwidth]{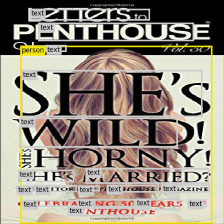}}
\framebox[0.22\columnwidth]{\includegraphics[width=0.22\columnwidth]{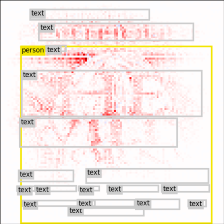}}
			
\framebox[0.44\columnwidth]{``Sports \& Outdoors''}
\framebox[0.06\columnwidth]{ }
\framebox[0.44\columnwidth]{``Romance''}
\end{center}
		
\begin{center}
\setlength\fboxsep{0pt}
\setlength\fboxrule{0pt}
\framebox[0.22\columnwidth]{\includegraphics[width=0.22\columnwidth]{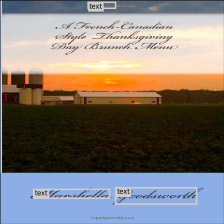}}
\framebox[0.22\columnwidth]{\includegraphics[width=0.22\columnwidth]{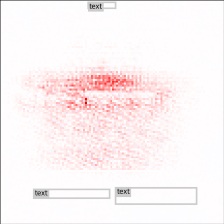}}
\framebox[0.06\columnwidth]{ }
\framebox[0.22\columnwidth]{\includegraphics[width=0.22\columnwidth]{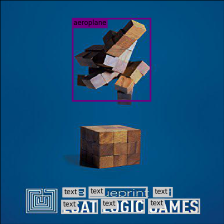}}
\framebox[0.22\columnwidth]{\includegraphics[width=0.22\columnwidth]{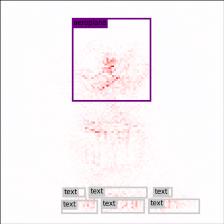}}
			
\framebox[0.44\columnwidth]{``Cookbooks,Food \& Wine''}
\framebox[0.06\columnwidth]{ }
\framebox[0.44\columnwidth]{``Test Preparation''}
\end{center}
\caption{\label{mis}  ``Misclassified'' book covers with correct genre names written below each book cover.}
\end{figure}

Such analysis helped us to find design elements specific to the presented genres. To get more familiar with design, we also presented some cases where the network was not able to correctly classify the genre. Figure~\ref{mis} shows some of these misclassifications, mainly from the presented genres. The correct genre names are written below the image. 
From the analysis presented above, one can easily decode the reason behind their misclassification because the designs on these book covers are not aligned with their genres which makes it obvious for network to mis-classify. Here, cover from "Sports \& Outdoors" contains birds, "Romance" cover contains text, "Cookbooks, Food \& Wine Books" contain no food picture and "Test Preparation" cover is also without any significant feature. LRP justifies all these covers misclassification by highlighting these mentioned objects contributing towards the "other" class in one-vs-others.

\section{How CNNs Understand Book Cover Design: Quantitative Analysis}	
\label{ssd}

\subsection{Experiment Setup}
In order to quantitatively analyze LRP, we propose using SSD to detect objects and EAST to detect text within the book cover images. 
We then use LRP to compare the relevance of objects bound by the detection methods. 
The SSD was trained on the \textit{2012 PASCAL Visual Object Classes}~(VOC) Challenge dataset~\cite{Everingham10}. 
The VOC dataset contains 20 classes, including "person," six animal classes, eight vehicle classes, and seven indoor object classes.
While SSD trained with VOC is intended for natural scene images, it can be used with book cover images because book covers often contain many of the shared classes, such as "person" and "car." 
Similarly, EAST was trained on the \textit{2015 ICDAR Robust Reading Competition} dataset~\cite{karatzas2015icdar} meant for scene text detection. 
Despite being trained for scene text, shown in Fig.~\ref{bookcovers}, EAST performs remarkably well on book covers for detecting text. 

To extract object and text bounding box information, the book covers were prepared by scaling the images to $512\times512$ pixels by 3 color channels. 
It is important to note that the images used for SSD and EAST were larger than the images used by the CNN used for genre classification. 
This is due to the detection methods being much more effective at the higher resolution. 
To accommodate this, the bounding boxes were scaled post detection and projected onto the LRP heatmaps. 
The relevance of an object $R_{\mathrm{obj}}$ is calculated using the sum of the relevance within the bounding box, or
\begin{equation}
    \label{eq:bbox}
    R_{\mathrm{obj}} = \sum_{(n,m)\in\mathcal{B}}{R_{(n,m)}},
\end{equation}
where $R_{(n,m)}$ is the relevance at pixel coordinates $(n,m)$ within bounding box $\mathcal{B}$.

\subsection{LRP with Object Detection} 
	
\begin{figure}
\centering
\includegraphics[width=1\columnwidth,trim={0cm 0cm 0cm 0cm},clip]{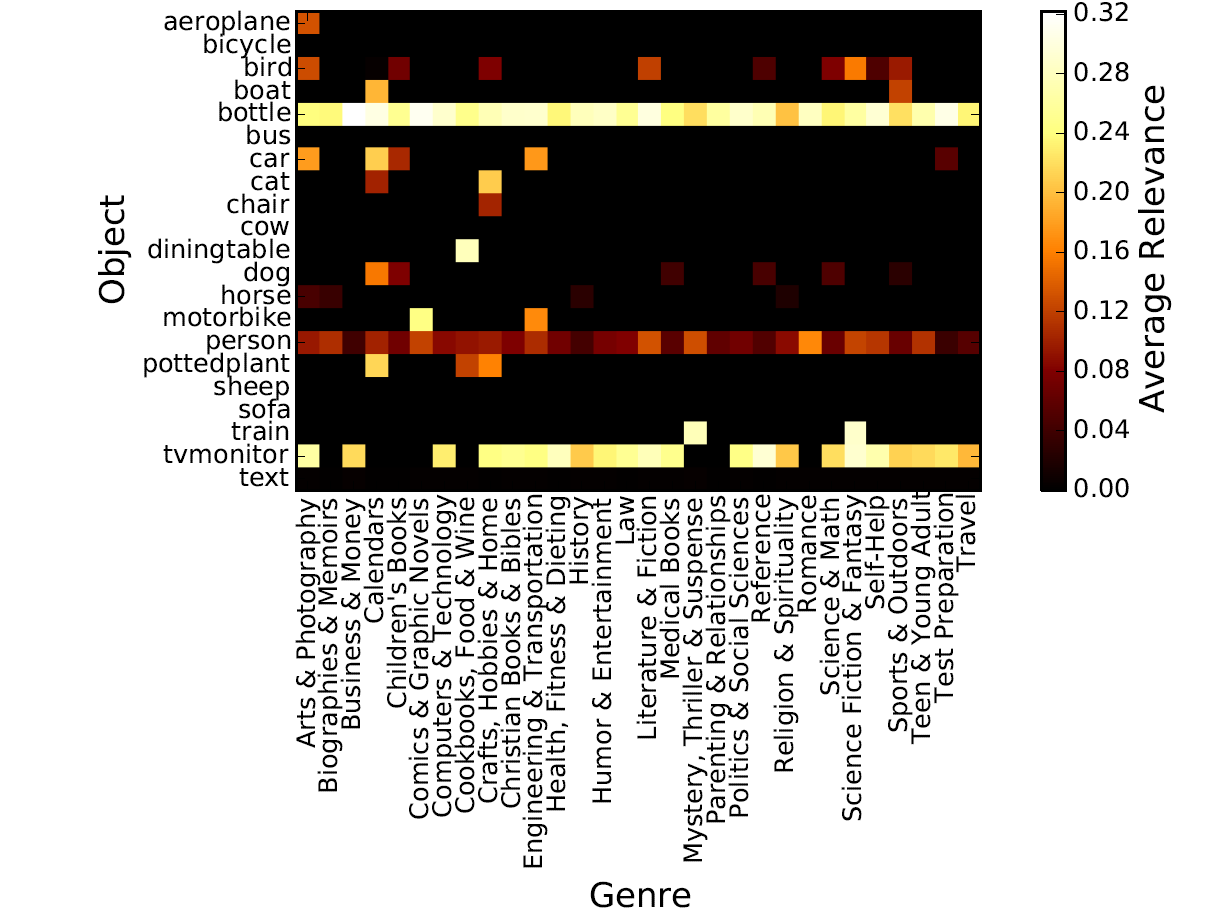}
\caption{\label{fig:objclass} Average object-wise relevance for text detected by EAST and each object class detected by SSD for each book genre. Only object-genre combinations with five or more data points are shown.}
\end{figure}

A macro view of the genres can be seen by viewing the average relevance of object classes. 
Figure~\ref{fig:objclass} illustrates the average object-wise relevance of each object class as detected by SSD and EAST for each book genre using the test set book cover images. 
It should be noted that detected objects such as "bottle" and "tvmonitor" were overfit to certain book cover images because many books have plain covers which resemble bottle labels or televisions. 
However, this does not mean that the information is useless. 
For example, from Fig.~\ref{fig:objclass}, "bottle" is more relevant for reference and nonfiction genres where plain covers are common.

\begin{figure}
\centering
\includegraphics[width=.9\columnwidth]{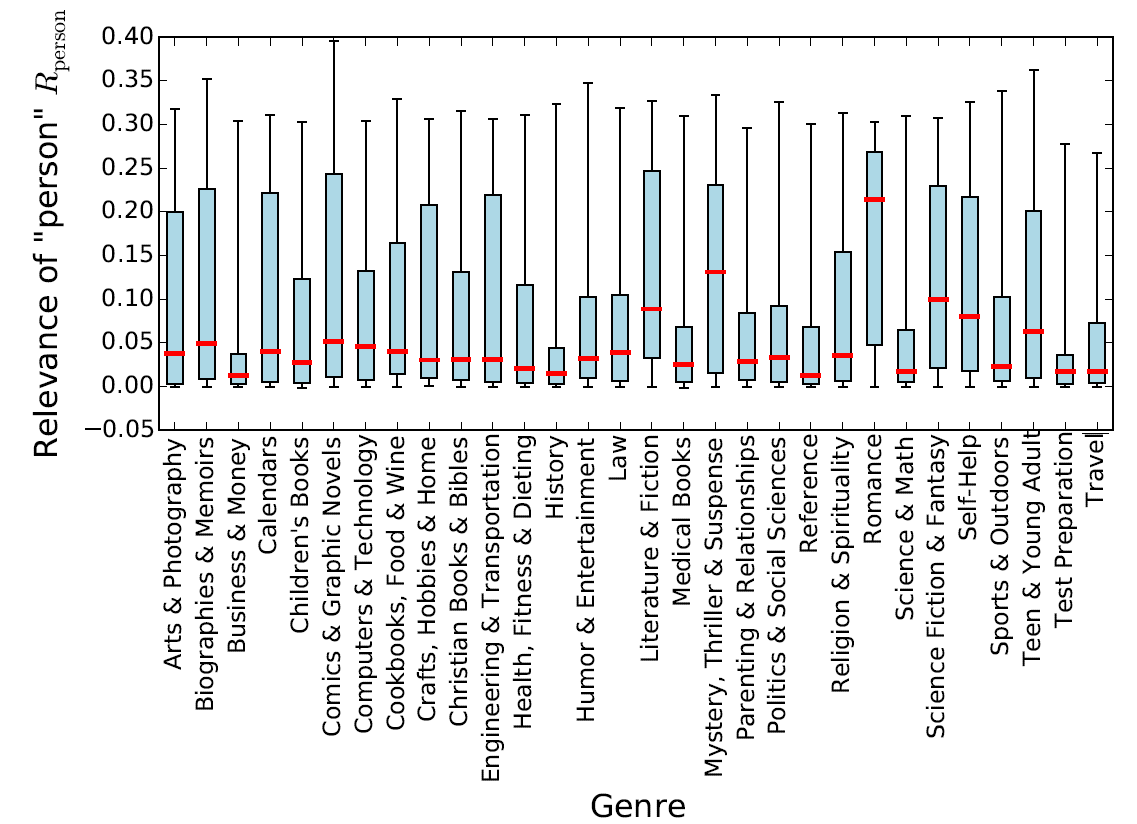}
\caption{\label{fig:person} Box plot of relevance of "person" $R_{\mathrm{person}}$ for each genre. The boxes represent the first through third quartile and the mean is in red. The whiskers mark the minimum and maximum datum.}
\end{figure}

In addition, by examining the distribution of the $R_{\mathrm{obj}}$ of specific object classes, such as "person," it is possible to create associations between genres and detected objects. 
For example, the relevance of "person" $R_{\mathrm{person}}$ for each genre is shown in Fig.~\ref{fig:person}.
The figure demonstrates that detected "person"s within certain genres are more relevant than other genres. 
For instance, the genres of "Romance" and "Mystery, Thriller \& Suspense" put a high average relevance in "person." 
This indicates that "person" is important for the CNNs of those categories. 
In addition, mentioned in Section~\ref{sec:testprep} and shown in Fig.~\ref{bookcovers}~(f), people are common in "Test Preparation" but are not necessarily relevant. 
This is supported by Fig.~\ref{fig:person} which indicates that on average, "person" has very little relevance.
Distributions for the other object classes are provided in the supplemental material.

\subsection{LRP with Text Detection}

\begin{figure}
\centering
\includegraphics[width=.9\columnwidth]{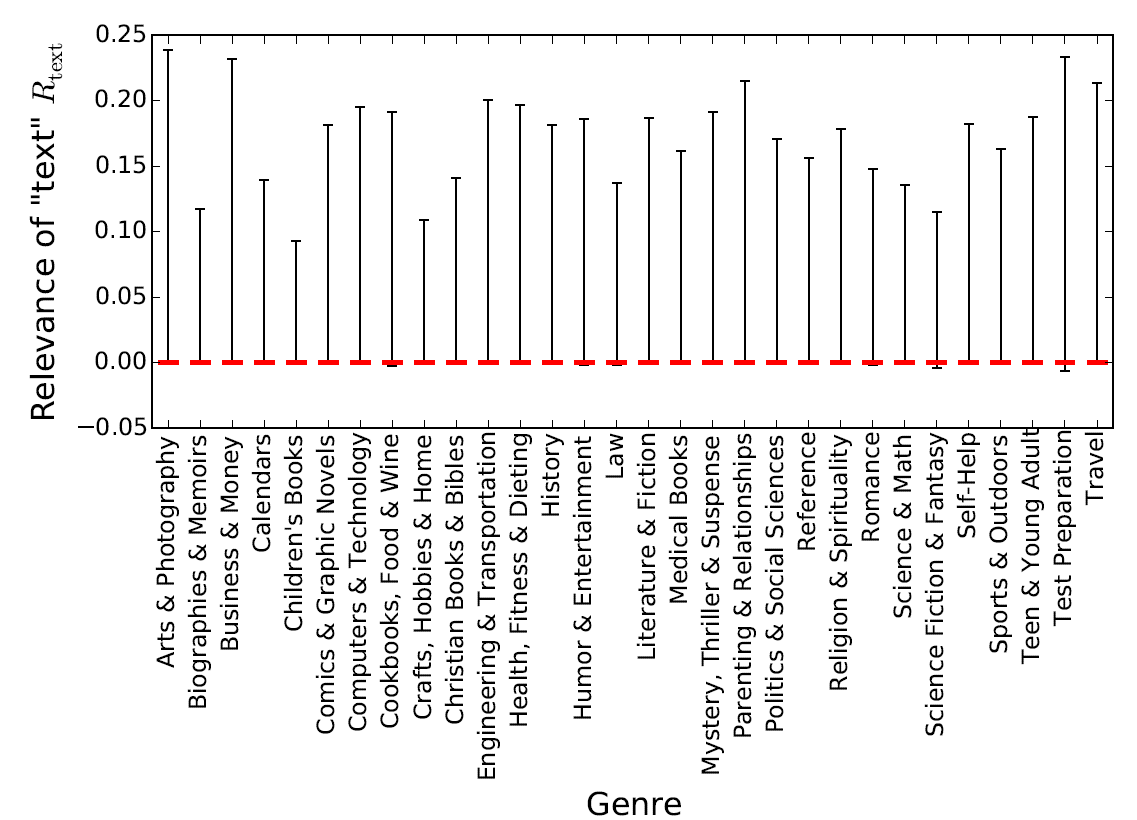}
\caption{\label{fig:text} Box plot of relevance of "text" $R_{\mathrm{text}}$ for each genre. The boxes represent the first through third quartile and the mean is in red. The whiskers mark the minimum and maximum datum.}
\end{figure}

Figure~\ref{fig:objclass} also reveals that the average relevance of text is low. 
The reasoning behind this phenomenon can be explained by Fig.~\ref{fig:text}. 
The figure shows that the majority of the detected text boxes have a very small relevance $R_{\mathrm{text}}$, but there are some text boxes have a higher relevance. 
For most genres, the title text contains a significant amount of relevance determined by LRP, but the small descriptive text carries very little relevance. 
Figure~\ref{bookcovers}~(f) in particular demonstrates this with the large title text having a high relevance and much of the smaller descriptive text having near zero relevance. 

\section{Conclusion}
\label{con}
In this paper, we presented importance of design in book covers belonging to a specific genre. The application of LRP on the book cover dataset showed genre specific book cover features. The method described most relevant parts of input book cover contributing towards a genre prediction by CNN. We also presented quantitative analysis of LRP using an object detection method, SSD, and a text detection method, EAST. The analysis further demonstrates that genre classification heavily relies on specific objects for each genres.

\section{ACKNOWLEDGEMENT}
This research was partially supported by MEXT-Japan (Grant No.J17H06100).

\bibliographystyle{IEEEtran}
\bibliography{book_shjo}
	
\end{document}